\documentclass[10pt,twocolumn,letterpaper]{article}

\usepackage{cvpr}              
\definecolor{cvprblue}{rgb}{0.21,0.49,0.74}
\usepackage[pagebackref,breaklinks,colorlinks,allcolors=cvprblue]{hyperref}
\usepackage[table]{xcolor}
\usepackage{booktabs}
\usepackage{multirow}
\usepackage{xcolor}
\usepackage{colortbl}
\usepackage{makecell}
\usepackage[most]{tcolorbox}
\definecolor{Gray}{gray}{0.95}

\def\paperID{xxx}
\def\confName{CVPR}
\def\confYear{2026}

\title{See Less, See Right: Bi-directional Perceptual Shaping For Multimodal Reasoning}

\author{
  Shuoshuo Zhang$^{\phi\pi}$\thanks{Equal contribution. Work done during internship at Microsoft.} \quad
  Yizhen Zhang$^{\phi}$\footnotemark[1] \quad
  Jingjing Fu$^{\pi}$ \quad
  Lei Song$^{\pi}$ \\
  Jiang Bian$^{\pi}$ \quad
  Yujiu Yang$^{\phi}$\thanks{Corresponding author.} \quad
  Rui Wang$^{\pi}$\footnotemark[2] \\
  $^{\pi}$ Microsoft Research Asia \quad
  $^{\phi}$ Tsinghua University \\
  {\tt\small zss24@mails.tsinghua.edu.cn, yang.yujiu@sz.tsinghua.edu.cn, ruiwa@microsoft.com}
}

\begin{document}
\maketitle

\begin{abstract}
Large vision–language models (VLMs) often benefit from intermediate visual cues, either injected via external tools or generated as latent visual tokens during reasoning, but these mechanisms still overlook fine-grained visual evidence (e.g., polylines in charts), generalize poorly across domains, and incur high inference-time cost. In this paper, we propose \textbf{Bi}-directional \textbf{P}erceptual \textbf{S}haping (\textbf{BiPS}), which transforms question-conditioned masked views into bidirectional where-to-look signals that shape perception during training. BiPS first applies a KL-consistency constraint between the original image and an evidence-preserving view that keeps only question-relevant regions, encouraging coarse but complete coverage of supporting pixels. It then applies a KL-separation constraint between the original and an evidence-ablated view where critical pixels are masked so the image no longer supports the original answer, discouraging text-only shortcuts (i.e., answering from text alone) and enforcing fine-grained visual reliance. Across eight benchmarks, BiPS boosts Qwen2.5-VL-7B by 8.2\% on average and shows strong out-of-domain generalization to unseen datasets and image types.
Code is available at~\url{https://github.com/zss02/BiPS}.
\end{abstract}    
\section{Introduction}
\label{sec:intro}

Large vision–language models (VLMs) are increasingly serving as a unified interface for both visual and language-based reasoning~\cite{bai2025qwen2,chen2024internvl}. Among real-world applications, visual question answering (VQA) is a widely deployed, high-impact task: a system must parse a natural-language query, localize the pertinent visual evidence, and produce an answer whose reasoning remains tied to that evidence. Despite rapid progress, the perceptual capability of VLMs, including the identification, localization, and accurate reading of fine-grained visual cues is still a bottleneck~\cite{liu2025perception,sim2025can,bigverdi2025perception}. If perception slips, downstream reasoning can rely on incomplete or misleading cues, yielding plausible but evidence-mismatched answers and degrading VQA performance.

\begin{figure}[t]
    \centering
    \includegraphics[width=\linewidth]{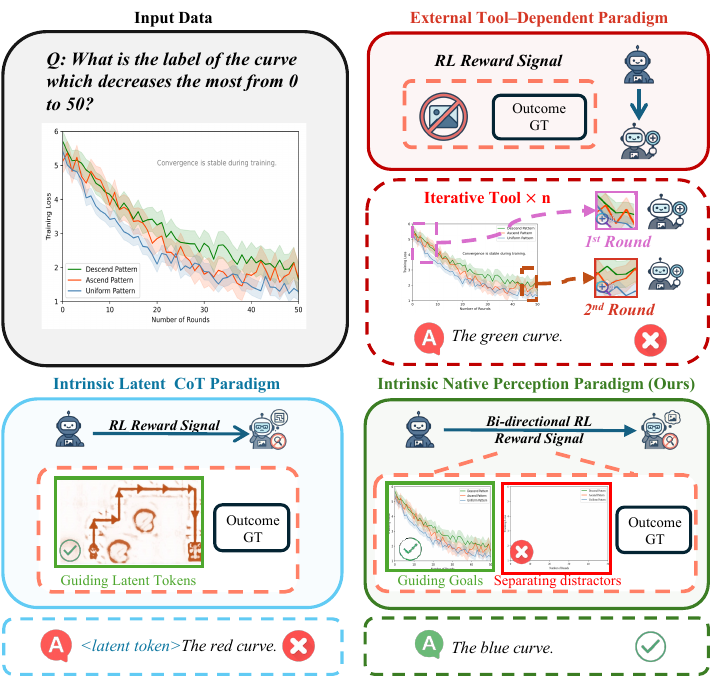}
    \caption{\textbf{Illustration of different paradigms.}
    Within each colored box, the top row illustrates the \textnormal{training} stage and the bottom row shows the \textnormal{inference} stage.
    Prior approaches are limited by shape-rigid inference-time tools and domain-specific solutions that generalize poorly.
    }
    \vspace{-0.2in}
    \label{fig:motivation}
\end{figure}

To mitigate this perceptual bottleneck, a complementary line of work augments VLMs with external visual tools (e.g., cropping, masking, segmentation) that produce evidence-focused intermediate visual cues at inference time~\cite{shao2024visual, wang2025multimodal,fu2025refocus,zhang2025pixelcraft,zheng2024instruction,hu2024visual,man2025argus}. Recent efforts further collect step-by-step “visual chain-of-thought’’ traces, where the model is guided by intermediate boxes, tool-use trajectories, or auxiliary images and then trained to reproduce these visual reasoning steps~\cite{shao2024visual,Chen2025MINTCoTEI,fu2025refocus,zheng2024instruction,yang2025machinementalimageryempower}. In practice, such mechanisms improve grounding and answer accuracy across tasks such as chart understanding, image-based math problems, and natural-image VQA with sparse cues.

Despite their utility, these approaches face three practical limitations. (i) Shape rigidity. Focused regions are typically rectangular crops or coarse masks, which miss irregular or fragmented evidence, such as thin polylines and intersections in charts, lesion contours in medical images, or nonconvex polygons in geometry diagrams. (ii) Scenario-specific solutions. Both custom tools and training pipelines that teach models to emit task-tailored latent visual tokens at inference time are tightly coupled to particular layouts or domains, limiting generalization. (iii) Inference-time overhead. Whether implemented via external tools or learned visual hints (e.g., boxes, masks, latent visual tokens), generating intermediate cues at inference introduces extra steps and computation and increases the risk of cascading errors. As illustrated in Fig.~\ref{fig:motivation}, we take a different route: programmatically generating perfect, ground-truth visual cues not as inference-time crutches, but as training signals. Rather than teaching the model to output specific visual cues, we use these cues to shape the model's internal policy, biasing it toward grounding its answers in visually supported content.

In this paper, we propose \textbf{Bi}-directional \textbf{P}erceptual \textbf{S}haping (\textbf{BiPS}), a two-stage training-time approach integrated into the Group Relative Policy Optimization (GRPO) framework~\cite{shao2024grpo},
which shapes a VLM’s perception by pulling predictions toward an evidence-preserving view and pushing them away from an evidence-ablated view. BiPS first builds a question-conditioned evidence-preserving view that keeps regions needed to answer the query while masking distractors, and applies a consistency constraint based on the Kullback–Leibler (KL) divergence to align the model’s predictions on this view with those on the original image, encouraging coarse but complete coverage of supporting pixels. It then constructs a complementary evidence-ablated view that finely removes critical pixels so the answer changes or becomes unanswerable, and adds a KL-based separation term that pushes predictions on the original and ablated views apart, discouraging text-only shortcuts (i.e., answering from text alone) and enforcing fine-grained visual reliance. This bidirectional shaping yields visually grounded decisions while requiring no extra annotations or customized parsers at inference.

Realizing BiPS requires precise question-conditioned supervision in the form of paired views, including an evidence-preserving view and an evidence-ablated view for each image and question pair. Naive pixel-level masking or cropping is common, but it remains coarse and shape-constrained. Complex, multi-panel charts naturally carry the fine-grained evidence (e.g., thin polylines, layered marks, and small symbols), making charts a rich source of training signals. We therefore build a programmatic data pipeline for chart data that generates the required evidence-preserving and evidence-ablated views. In this work, we instantiate the pipeline using ECD~\cite{yang2025ecd10k}, a synthetic corpus of complex multi-panel figures paired with executable rendering code. Because each figure is generated by code, every object (marks, layers, axes, legends) has explicit provenance, which enables exact edits to synthesize the two complementary views. This pipeline yields 13K synthetic training examples. Fine-tuned solely on this set, BiPS already generalizes well: across eight benchmarks spanning real-world figure datasets (e.g., CharXiv~\cite{wang2024charxiv}, ChartQAPro~\cite{masry2025chartqapro}) and out-of-domain general VQA (e.g., MathVista~\cite{lu2023mathvista}, MMStar~\cite{chen2024mmstar}), average accuracy improves by +7.3\% over the base model (Qwen2.5-VL-7B~\cite{bai2025qwen2}). Adding 39k math-specialized examples with standard GRPO further increases the average gain to 8.2\%.

The contributions are summarized as follows:
\begin{itemize}
    \item We turn inference-time visual cues into training signals that internalize perception. BiPS applies a KL consistency term toward an evidence-preserving view and a KL separation term from an evidence-ablated view. These signals teach the model to capture visual evidence, including fine and irregular details, yielding evidence-faithful predictions without test-time overhead.
    \item We build a programmatic data construction pipeline that uses executable chart scripts to synthesize precise paired views (preserve and ablate) without human labeling. Fine-tuned solely on chart-derived cues, BiPS extends well to real-world figures and general VQA, indicating strong out-of-domain generalization.
    \item Experiments show that BiPS delivers substantial, consistent performance gains, with strong cross-domain generalization and high data efficiency. 
    Trained on only 13K chart samples,
    it boosts Qwen2.5-VL-7B by 7.3\% on average across eight widely used chart and general VQA benchmarks (e.g., CharXiv, MathVista, MMStar),
    surpassing specialized models trained on vastly larger datasets.
    Adding 39K math-focused samples with GRPO further increases the average improvement to 8.2\%.
\end{itemize}
\section{Related Works}

\begin{figure*}[t]
    \centering
    \includegraphics[width=\textwidth]{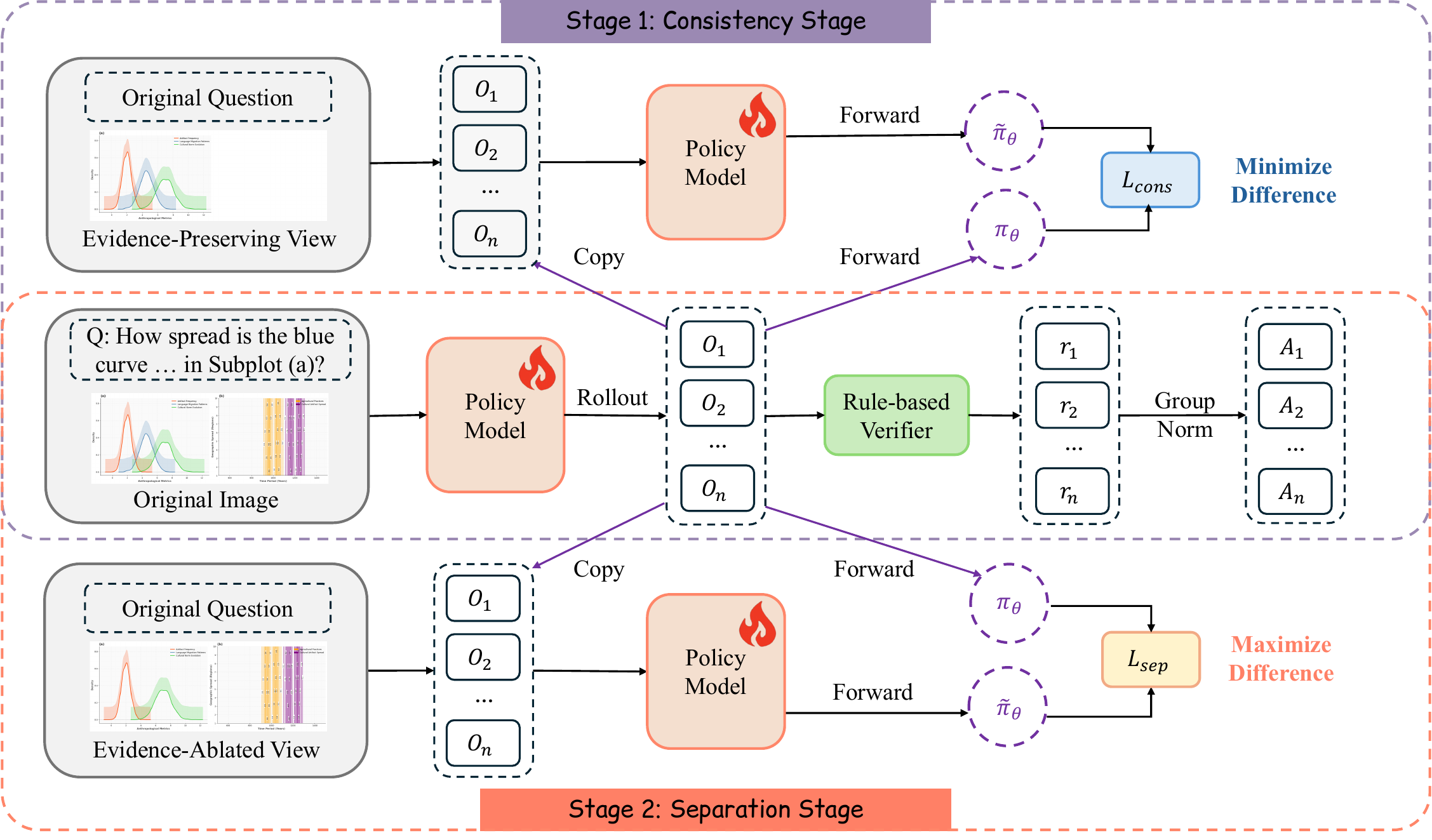}
    \caption{\textbf{Overview of the Bi-directional Perceptual Shaping (BiPS) framework.}
    BiPS employs a two-stage training curriculum built on the GRPO framework. Stage 1 (Consistency Stage) minimizes the KL-divergence ($L_{cons}$) between the original policy ($\pi_{\theta}$) and the policy on an evidence-preserving view ($\tilde{\pi}_{\theta}$). Stage 2 (Separation Stage) maximizes the KL-divergence ($L_{sep}$) between the original policy ($\pi_{\theta}$) and the policy on an evidence-ablated view, forcing the model to ground its reasoning in visual evidence.
    }
    \vspace{-10pt}
    \label{fig:framework}
\end{figure*}

\subsection{Multimodal Reasoning}

Recent advances in reinforcement learning (RL) for language models have sparked growing efforts to extend RL to multimodal reasoning~\cite{Luo2025Ursa,zhang2025generative}. Recent studies apply RL to strengthen visual understanding or reasoning across domains, including single-image~\cite{meng2025mmeureka,yang2025r1onevision,huang2025visionr1,Chen2025revisualr1,wang2025vlrethinker} reasoning for spatial perception, multi-image reasoning~\cite{zhang2025perl} for cross-scene consistency, video reasoning ~\cite{Feng2025VideoR1RV} for temporal causality, and chart reasoning ~\cite{chen2025chartr1} for quantitative alignment.

Complementary to advances in language reasoning, a growing body of work improves VLMs by injecting intermediate visual cues at inference time. Typical strategies highlight question-relevant content via bounding boxes~\cite{shao2024visual}, question-conditioned crops~\cite{man2025argus}, or masks that suppress distractors~\cite{zheng2024instruction}, biasing predictions toward the appropriate visual evidence. Building on this idea, subsequent methods design task-aware tools or specialized modules for tighter evidence localization, such as chart-specific annotation tools~\cite{fu2025refocus,zhang2025pixelcraft} and visual sketchpad~\cite{hu2024visual}. Pushing this paradigm further, recent studies supervise models with labeled evidence and train them to reproduce coordinates, masks, or tool selections so that downstream answers can be conditioned on the predicted evidence~\cite{fu2025refocus, shao2024visual,zheng2024instruction,su2025openthinkimg,wu2025vtool,sarch2025grounded,Zheng2025DeepEyesI,Su2025PixelReasoner,Chen2025MINTCoTEI}. However, these designs inherit several limitations: rectangular or coarse masks struggle with fine-grained or irregular structures, task-specific engineering reduces generalization, and multi-step pipelines introduce nontrivial inference overhead.

A recent work removes the need for explicit intermediate images by encouraging models to reason through latent internal processes~\cite{yang2025machinementalimageryempower}, but such approach remain confined to specific tasks. ChiP~\cite{fu2025chip} and PAPO~\cite{wang2025papo} inject Gaussian noise or random masks as negative perturbations that penalize reliance on incorrect images to avoid visual hallucinations and text-only shortcuts, yet these methods overlook informative detailed signals inside the figure.

\subsection{Chart Understanding And Reasoning}

Unlike general multimodal reasoning on natural scenes~\cite{Fu2023MMEAC} or geometric diagrams~\cite{Lu2021InterGPSIG,lu2023mathvista,wang2024mathvision}, chart understanding targets structured quantitative graphics where numerical relations are encoded by axes and marks, demanding precise value perception and visual–numerical correspondence. Early chart-domain studies mainly focused on low-level perception tasks such as chart element detection and text extraction~\cite{Methani2019PlotQARO,Masry2022ChartQAAB}, while recent benchmarks shift toward reasoning-centric tasks~\cite{wang2024charxiv,tang2025chartmuseum,masry2025chartqapro} that require interpreting implicit patterns, making comparisons, and executing multi-step computations over visualized data. To enhance such reasoning, subsequent studies convert charts into structured symbolic programs or code for executable reasoning~\cite{Zhang2024TinyChartEC,Jia2025ChartReasonerCM,Zhao2025ChartCoderAM,yang2024chartmimic}, while others explore multimodal feedback, reflective learning, and reinforcement-based optimization~\cite{Huang2025ChartSketcherRW,masry2025bigcharts,chen2025chartr1} to improve reliability. Nonetheless, chart reasoning still faces challenges such as limited chart-to-code accuracy~\cite{Masry2022ChartQAAB,Xu2023ChartBenchAB}, weak generalization to diverse layouts, and difficulty handling high-frequency curve fluctuations or subtle oscillatory patterns, which often cause quantitative misalignment. Addressing these challenges demands finer-grained perception and robust visual reasoning.

\label{sec:formatting}

\section{Method}

To address the core perceptual failures of VLMs, namely their tendency to be distracted by irrelevant visual information and their inability to focus on fine-grained evidence, we propose BiPS. 
As illustrated in Figure~\ref{fig:framework}, BiPS is implemented as a two-stage training curriculum consisting of a Consistency Stage followed by a Separation Stage.
The Consistency Stage guides the model to maintain consistent predictions when only evidence-relevant regions are retained, 
whereas the Separation Stage drives the model to diverge when the critical regions are removed.
Both stages are implemented as bidirectional KL-based objectives, optimized within the GRPO~\cite{shao2024grpo} framework. 

\subsection{Programmatic Data Construction Pipeline}
\label{sec:data_construction}
\begin{figure}[t]
    \centering
    \includegraphics[width=\linewidth]{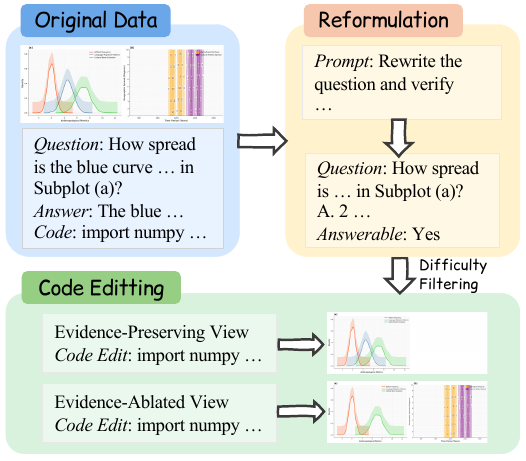}
    \caption{\textbf{Overview of our data generation pipeline.}
    This pipeline programmatically edits chart source code to generate the paired Evidence-Preserving ($I_\mathrm{{pres}}$) and Evidence-Ablated ($I_\mathrm{{abl}}$) views used for bi-directional training.
    }
    \vspace{-10pt}
    \label{fig:data_pipeline}
\end{figure}
To enable BiPS to learn evidence-aware grounding, we require training supervision that precisely distinguishes relevant from irrelevant visual content.
Such supervision takes the form of paired visual views---an evidence-preserving view that retains the regions needed to answer the question, and an evidence-ablated view where critical regions are removed.
While masking or cropping images is a common way to approximate such pairs, these operations are often coarse and shape-limited (e.g., rectangles), failing to capture fine-grained or irregular evidence.

We overcome this limitation by constructing the paired views programmatically in the code domain.
Our data generation pipeline builds upon ECD~\cite{yang2025ecd10k}, which provides complex, multi-panel charts along with their executable rendering code.
Charts serve as an ideal substrate for this purpose: their elements (marks, axes, legends) are semantically structured, fine-grained, and explicitly linked to the underlying code.
This enables editing at the code level rather than pixels, yielding semantically precise and fine-grained control over visual evidence.
As shown in Figure~\ref{fig:data_pipeline}, our pipeline consists of the following three main stages.

\paragraph{Question Reformulation and Validation.}
The original dataset's reasoning questions are open-ended and difficult to verify by predefined rules. 
Following a typical RLVR setting~\cite{shao2024grpo,meng2025mmeureka}, we refine this data by employing an auxiliary LLM arbitrator, GPT5-mini, to convert the original questions into a multiple-choice format. 
This arbitrator is provided with the chart's source code and metadata to ensure that the reformulated question remains answerable and that the ground-truth option is correct. 
This step provides verifiable supervision compatible with downstream RL training and ensures the overall quality of our base question set.

\paragraph{Difficulty Filtering.}
To focus the training on non-trivial examples, we filter out questions that are too ``easy'' for the base model (Qwen2.5-VL-7B-Instruct~\cite{bai2025qwen2}). We perform 8 rollouts for each validated question. 
Any question that the base model answers correctly in all rollouts is considered ``easy'' and is discarded from our training set, allowing our method to concentrate on more challenging reasoning tasks.

\paragraph{Code Editing and Counterpart Rendering.}
This is the core stage where we generate the paired visual counterparts required for BiPS training. 
For each filtered question $q$ and its corresponding chart-rendering code $C$, we employ the LLM arbitrator guided by structured prompts to identify and modify the relevant code components.

\begin{itemize}
    \item \textbf{Evidence-Preserving View ($I_\mathrm{{pres}}$):}
    To construct the minimal evidence view used in the Consistency Stage, 
    the arbitrator removes code segments unrelated to answering $q$ and executes the remaining script to render an image containing only the necessary visual elements.

    \item \textbf{Evidence-Ablated View ($I_\mathrm{{abl}}$):}
    To create the complementary view for Separation Stage,
    the arbitrator identifies code segments that provide the key evidence and removes them while keeping general contextual structures such as axes, legends, and layout. 
    The resulting image omits fine-grained cues yet preserves the global context.
\end{itemize}
This programmatic process produces a large-scale dataset of $(I, q, I_\mathrm{{pres}}, I_\mathrm{{abl}})$ tuples, 
providing semantically precise and well-aligned supervision for BiPS.
After this pipeline, we obtained a final set of 13K high-quality training samples (Detailed data statistics are available in Appendix).

\subsection{Bi-directional KL Constraints}
\label{sec:kl_constraint}
Our core methodology comprises two complementary KL constraints operating on the paired views constructed in Section~\ref{sec:data_construction}. 
The first constraint provides the primary positive guidance for evidence localization, while the second provides a negative-space regularization to ensure that this localization is robust and visually grounded.

\subsubsection{Focusing via Consistency}
To teach the model to ignore distractions and focus on the correct region, we use a Consistency constraint, which enforces that the model's policy on the full image $I$ should be consistent with its policy on the evidence-preserving view $I_{pres}$. 
We apply this constraint by minimizing the KL-divergence between these two distributions:
\begin{equation}
\label{eq:consistency}
\begin{split}
\mathcal{L}_{\text{cons}}
&=
\mathbb{E}_{(I,q,r)}\Big[
\mathbb{I}(r{=}1)\,
\min\Big(
c_{\text{cons}},\\
&\qquad\quad
\mathbb{D}_{\mathrm{KL}}\!\big(
\pi_{\theta}(\cdot\!\mid\! I,q)
\,\big\|\,
\operatorname{sg}\!\big[\tilde{\pi}_{\theta}(\cdot\!\mid\! I_{\text{pres}},q)\big]
\big)
\Big)
\Big].
\end{split}
\end{equation}
\noindent
Here $\pi_{\theta}$ is the model’s answer distribution; 
$\tilde{\pi}_{\theta}$ denotes a target distribution computed on the evidence-preserving view.
$\operatorname{sg}[\cdot]$ indicates stop-gradient so that the $I_{\text{pres}}$ branch serves as a fixed target; 
$\mathbb{I}(r{=}1)$ restricts supervision to validated correct samples; 
and $c_{\text{cons}}$ clips the KL term for stability.

The forward direction $\mathbb{D}_{\mathrm{KL}}\!\big(\pi_{\theta}\,\|\,\tilde{\pi}_{\theta}\big)$ pulls probability mass on $I$ toward evidence-supported answers on $I_{\text{pres}}$, encouraging the policy to base decisions on preserved evidence and treat extraneous regions as redundant.

\subsubsection{Robustness via Separation}
On its own, $\mathcal{L}_{\text{cons}}$ is insufficient as it is susceptible to ``shortcut learning''. A model may satisfy the consistency $\mathcal{L}_{\text{cons}}$ by exploiting surrounding text (OCR) or language priors to produce identical answers for the original and evidence-preserving images, without attending to fine-grained details (e.g., curves).

To ensure the model's focus is truly visually grounded, we introduce a complementary Separation constraint, which acts as a regularizer, forcing the model to learn that the visual signal is indispensable. It achieves this by enforcing that the model's policy on the full image $I$ must be divergent from its policy on the evidence-ablated View $I_\mathrm{{abl}}$. We maximize the KL-divergence between these distributions:
\begin{equation}
\label{eq:separation_max}
\begin{split}
\mathcal{L}_{\text{sep}}
&=
\mathbb{E}_{(I,q)}\Big[
\min\Big(
c_{\text{sep}},\\
&\qquad\quad
\mathbb{D}_{\mathrm{KL}}\!\big(
\pi_{\theta}(\cdot\!\mid\! I,q)
\,\big\|\,
\operatorname{sg}\!\big[\tilde{\pi}_{\theta}(\cdot\!\mid\! I_{\text{abl}},q)\big]
\big)
\Big)
\Big].
\end{split}
\end{equation}
\noindent where $c_{\text{sep}}$ is a clipping hyperparameter.
This objective penalizes similarity between the two policies until the divergence exceeds $c_{\text{sep}}$, breaking text-only shortcuts and promoting fine-grained grounding. 

\subsection{A Coarse-to-Fine Training Curriculum}
\label{sec:curriculum}
Optimizing $\mathcal{L}_{\text{cons}}$ (an attractive force) and $\mathcal{L}_{\text{sep}}$ (a repulsive force) simultaneously can be challenging due to potentially conflicting gradients. We therefore devise a two-stage curriculum that decouples these objectives.

\paragraph{Base GRPO Training Objective.}
Before detailing the stages, we define the base GRPO~\cite{shao2024grpo} objective, which extends Proximal Policy Optimization (PPO) by normalizing rewards across rollouts within the same group to stabilize training. The training objective is:
\begin{equation}
\label{eq:grpo_simple}
\begin{split}
\mathcal{L}_{\mathrm{GRPO}}
&=
-\mathbb{E}_{(I,q)}\Big[
\min\big(
r_t(\theta)A_t,\,
\mathrm{clip}(r_t(\theta),\\
&\qquad\quad
1-\epsilon,\,
1+\epsilon)A_t
\big)
-
\gamma D_{KL}\big(
\pi_\theta\|\pi_{\mathrm{ref}}
\big)
\Big].
\end{split}
\end{equation}
Here, $A_t$ denotes the group-relative advantage, $\epsilon$ is the clipping threshold, and $\gamma$ controls the KL penalty strength.

\paragraph{Stage 1: Consistency Stage.}
In the first stage, we train the model on the primary task of evidence localization using the consistency constraint:
\begin{equation}
\label{eq:stage1}
\mathcal{L}_{\text{Stage1}} = 
\mathcal{L}_{\mathrm{GRPO}} + 
\alpha\,\mathcal{L}_{\text{cons}}
\end{equation}
where $\alpha$ is the consistency constraint coefficient. 
This stage establishes the foundational, coarse-grained skill of what to focus on.

\paragraph{Stage 2: Separation Stage.}
Building on the Stage 1 checkpoint, we now introduce the separation constraint using a view that removes fine-grained visual evidence, ensuring the learned focus is robust and truly visually grounded: 
\begin{equation}
\label{eq:stage2}
\mathcal{L}_{\text{Stage2}} =
\mathcal{L}_{\mathrm{GRPO}} -
\beta\,\mathcal{L}_{\text{sep}}
\end{equation}
where $\beta$ is the separation constraint coefficient. 

This coarse-to-fine curriculum first applies the positive signal ($\mathcal{L}_{\text{cons}}$) and then the regularizer ($\mathcal{L}_{\text{sep}}$) to ensure the learned policy is both accurate and grounded. We demonstrate the superiority of this curriculum over joint training and the reversed order in our ablation studies (Sec.~\ref{sec:ablation}).
\section{Experiments}
\definecolor{closedblue}{RGB}{235,243,255}
\definecolor{opensourcepeach}{RGB}{255,244,232}
\definecolor{reasoningmint}{RGB}{230,247,237}
\definecolor{ourslavender}{RGB}{240,240,248}
\definecolor{deltarose}{RGB}{250,236,240}

\definecolor{SFTblue}{RGB}{31,98,178}
\definecolor{RLred}{RGB}{188,46,34}

\newcommand{\charticon}{\includegraphics[height=1.15em]{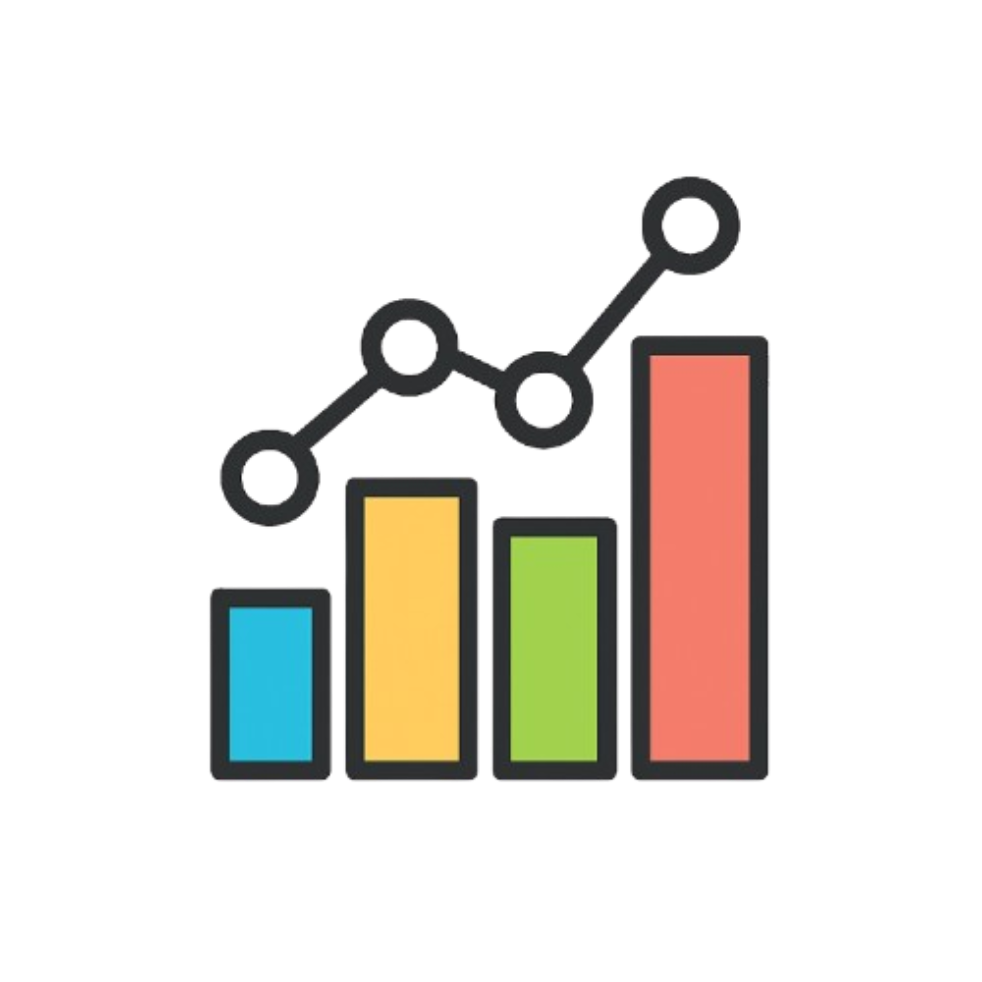}}
\newcommand{\mathicon}{\includegraphics[height=1.15em]{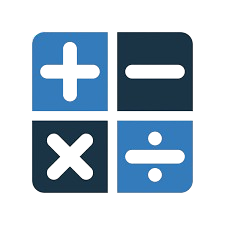}}

\newcommand{\inc}[1]{\textcolor{red}{\(\uparrow\,#1\)}}
\newcommand{\chart}[1]{\textcolor{SFTblue}{#1}}
\newcommand{\mc}[1]{\textcolor{RLred}{#1}}

\begin{table*}[t]
\centering
\footnotesize
\renewcommand{\arraystretch}{1.12}
\setlength{\tabcolsep}{2.7pt}
\caption{
\textbf{Evaluation on chart understanding and general perception \& reasoning benchmarks.}
Models with \charticon{} are chart-specialized; models with \mathicon{} are math-specialized; 
\textsuperscript{†} denotes our models. \textcolor{SFTblue}{Blue} numbers denote chart-related data, and \textcolor{RLred}{red} numbers denote math-related or perception-related data.
\emph{Avg.} is the arithmetic mean over available metrics. Models trained on \textbf{MathVision} (marked with $^{\star}$) omit their scores on MathVision.
Best scores are in \textbf{bold}, and second-best scores are \underline{underlined}. Baselines are reproduced using VLMEvalKit to ensure strict fairness.
}
\vspace{-8pt}
\begin{tabular}{l|c|cccc|cccc|c}
\toprule
\multirow{2}{*}{\textbf{Model}} &
\multirow{2}{*}{\textbf{Data}} &
\multicolumn{4}{c|}{\textbf{Chart Understanding and Reasoning}} &
\multicolumn{4}{c|}{\textbf{General Perception and Reasoning}} &
\multirow{2}{*}{\textbf{Avg.}} \\
\cmidrule(lr){3-6} \cmidrule(lr){7-10}
 &  & CharXiv & ChartQAPro & ChartMuseum & Evochart & MathVista & MathVision & MathVerse-VO & MMStar &  \\
\midrule
\rowcolor{closedblue}
\multicolumn{11}{l}{\textbf{Closed-Source Models}} \\
GPT-4o & -  & 47.1 & 37.7 & 42.2 & 49.8 & 63.8 & 31.2 & 40.6  & 65.1  & 47.2 \\
Claude-3.7-Sonnet & - & 64.2 & -- & 60.3 & 70.7 & 74.5 & 58.6 & 52.0  & 68.8 & -- \\
\midrule
\rowcolor{opensourcepeach}
\multicolumn{11}{l}{\textbf{Open-Source General Models}} \\
Qwen2.5-VL-7B & -  & 42.5 & 36.6 & 26.8 & 52.0 & 68.2 & 25.2 & 41.1 & 62.1 & 44.3 \\
InternVL3-8B  & -  & 37.6 & 36.9 & 28.2 & 55.0 & 70.4 & 26.3 & 33.9 & 68.2 & 44.6 \\
\midrule
\rowcolor{reasoningmint}
\multicolumn{11}{l}{\textbf{Multimodal Reasoning Models}} \\
ChartLlama-13B~\charticon & \chart{160K}  & 14.2 & -- & -- & 9.5 & -- & -- & -- & -- & -- \\
ChartGemma-3B~\charticon & \chart{163K}  & 12.5 & 6.8 & 12.2 & 30.6 & -- & -- & -- & -- & -- \\
TinyChart-3B~\charticon & \chart{1.3M}  & 8.3 & 13.3 & 12.5 & 25.5 & -- & -- & -- & -- & -- \\
R1-OneVision-7B~\charticon~\mathicon & \chart{67K} + \mc{98K} & 33.8 & 36.1 & 27.2 & 35.1 & 64.1 & \textbf{29.9} & 40.0 & 52.2 & 39.8 \\
Vision-R1-7B$^{\star}$~\charticon~\mathicon & \chart{73K} + \mc{137K}& 42.5 & 39.6 & 28.5 & 59.9 & 73.2 & - & \textbf{47.7} & 64.8 & -- \\
DeepEyes-7B ~\charticon
~\mathicon & \chart{14K} + \mc{33K} & 42.9 & 38.1 & 28.1 & 65.6 & 70.8 & 26.5 & 44.9 & 63.0 & 47.5 \\
BigCharts-R1-7B~\charticon & \chart{1.7M} & 41.3 & - & - & - & - & - & - & -  & -- \\
Chart-R1-7B~\charticon & \chart{258K} & 46.2 & 44.0 & 31.7 & 64.7 & 67.5 & 20.6 & 28.1 & 61.1 & 45.5 \\
GRPO~\charticon
~\mathicon & \chart{13K} + \mc{39K} & 45.4 & 50.2 & 32.9 & 68.0 & 74.3 & 27.3 & 42.6 & 64.6 & 50.7 \\
\rowcolor{ourslavender}
\textbf{BiPS-Chart-7B\textsuperscript{†}~\charticon} & \chart{13K} & \underline{49.4} & \textbf{51.9} & \underline{33.5} & \underline{68.2} & \underline{73.5} & 27.2 & 44.4 & \underline{64.9} & \underline{51.6} \\
\rowcolor{deltarose}
\emph{$\Delta$ over base model} &   & \inc{6.9} & \inc{15.3} & \inc{6.7} & \inc{16.2} & \inc{5.3} & \inc{2.0} & \inc{3.3} & \inc{2.8} & \textbf{\inc{7.3}} \\
\rowcolor{ourslavender}
\textbf{BiPS-General-7B\textsuperscript{†}~\charticon~\mathicon} & \chart{13K}+\mc{39K} & \textbf{50.6} & \underline{51.8} & \textbf{34.0} & \textbf{68.7}& \textbf{75.0} & \underline{28.6} & \underline{45.3} & \textbf{65.7} & \textbf{52.5} \\
\rowcolor{deltarose}
\emph{$\Delta$ over base model} &   & \inc{8.1} & \inc{15.2} & \inc{7.2} & \inc{16.7} & \inc{6.8} & \inc{3.4} & \inc{4.2} & \inc{3.6} & \textbf{\inc{8.2}} \\
\bottomrule
\end{tabular}
\label{tab:main_results}
\end{table*}

\subsection{Experiment Setting}
\paragraph{Implementation.}
\label{sec:impl_main}
We use Qwen2.5-VL-7B~\cite{bai2025qwen2} as the base model. 
Stage~1 is trained for 5 epochs on 7K samples containing evidence-preserving views $I_{\mathrm{pres}}$. 
Subsequently, Stage~2 trains for 3 epochs on 13K samples including evidence-ablated views $I_{\mathrm{abl}}$, producing our BiPS-Chart model.
To further enhance general reasoning capabilities, we re-optimize the Stage~2 checkpoint for 3 epochs on 39K samples from ViRL39k~\cite{wang2025vlrethinker} using standard GRPO, yielding the final BiPS-General. 
All models are optimized with AdamW ($lr=1\times10^{-6}$) on 8$\times$H100 GPUs. 
We set the constraint coefficients to $\alpha=0.01$ and $\beta=0.02$, and the clipping thresholds to $c_{\text{cons}}=1.0$ and $c_{\text{sep}}=0.2$.

\paragraph{Benchmark.}
We evaluate our models on a comprehensive suite of benchmarks spanning two categories. 
To assess chart understanding and reasoning capabilities, we report scores on CharXiv~\cite{wang2024charxiv}, ChartQAPro~\cite{masry2025chartqapro}, ChartMuseum~\cite{tang2025chartmuseum}, Evochart~\cite{huang2025evochart} and
ECD-Bench~\cite{yang2025ecd10k}.
For general perception and reasoning, We use MathVista~\cite{lu2023mathvista}, MathVision~\cite{wang2024mathvision}, MathVerse-VO~\cite{zhang2024mathverse}, and MMStar~\cite{chen2024mmstar}. 

\paragraph{Baseline.}
We evaluate our method against diverse baselines, including closed-source models like GPT-4o~\cite{hurst2024gpt} and Claude-3.7-Sonnet as high-level SOTA references. 
Our open-source comparisons include general models such as InternVL3-8B~\cite{zhu2025internvl3} and Qwen2.5-VL-7B~\cite{bai2025qwen2}. 
We also compare against specialized reasoning models, including chart-focused systems (ChartLlama~\cite{han2023chartllama}, ChartGemma~\cite{masry2025chartgemma}, Chart-R1~\cite{chen2025chartr1}, BigCharts-R1~\cite{masry2025bigcharts}) and broader multimodal reasoners (R1-OneVision~\cite{yang2025r1onevision}, Vision-R1~\cite{huang2025visionr1}, DeepEyes~\cite{Zheng2025DeepEyesI}) that handle both chart and math reasoning.

\subsection{Main Results}
Table~\ref{tab:main_results} reports performance across a diverse set of chart understanding and general multimodal reasoning benchmarks. 
Overall, our approach delivers substantial improvements over the baseline Qwen2.5-VL-7B, boosting the average score by +7.3 points (from 44.3 to 51.6) with BiPS-Chart-7B and by a total of +8.2 points (to 52.5) with BiPS-General-7B. 
This reflects clear and consistent gains in both chart-centric and general reasoning performance.

BiPS-Chart-7B first illustrates the power of our method, achieving significant gains on both chart-specific and general out-of-domain benchmarks.
Trained on only 13K chart samples, it delivers substantial gains on challenging chart reasoning benchmarks.
On ChartXiv, performance increases from 42.5 (Qwen2.5-VL-7B) to 49.4 (+6.9), and on Evochart from 52.0 to 68.2 (+16.2).
Crucially, BiPS-Chart-7B also exhibits strong OOD generalization, boosting performance on unseen general reasoning tasks such as MathVista (+5.3) and MMStar (+2.8).
These improvements surpass other chart-specialized models that rely on far larger training sets, including TinyChart-3B, BigCharts-R1-7B, and Chart-R1-7B, which are trained on hundreds of thousands to millions of chart examples but achieve lower scores on these key benchmarks. 
This comparison underscores that BiPS, by enhancing the model's core visual perception capabilities, enables a more data-efficient acquisition of chart reasoning and promotes robust generalization across diverse chart types.

Building on these gains, BiPS-General-7B incorporates 39K math-focused samples to further enhance the model's reasoning capacity. This integration leads to consistent improvements across chart and general benchmarks. 
BiPS-General-7B reaches 50.6 on ChartXiv, 68.7 on Evochart, 75.0 on MathVista, and 65.7 on MMStar, surpassing both the baseline and the chart-only BiPS-Chart-7B model. This demonstrates that the introduction of math-specific data further compounds these gains by explicitly strengthening the model's abstract and numerical reasoning abilities.

To verify that the improvements stem from our specific perceptual shaping pipeline rather than merely applying RL to the data, we compare BiPS-General-7B against a baseline trained with standard GRPO. 
This baseline is fine-tuned on the exact same combined dataset (programmatic samples + ViRL39k) but treats all data uniformly without the proposed two-stage perceptual shaping curriculum.
While standard GRPO yields substantial gains over the base model, \textbf{BiPS-General} consistently outperforms it across all benchmarks.
Notably, on complex chart reasoning tasks like CharXiv, our method surpasses the standard GRPO baseline by a significant margin (\textbf{+5.2}).
This performance gap highlights a critical insight: simply optimizing reasoning via RL is insufficient if the underlying visual grounding is flawed.
By explicitly shaping the model's perception through our programmatic curriculum, BiPS ensures that the RL process operates on high-fidelity visual signals, thereby amplifying the effectiveness of the optimization.

\subsection{Ablation Study}
\label{sec:ablation}

\paragraph{Impact of Bi-directional KL Constraints.}
\begin{table}[t]
\centering
\caption{
\textbf{Ablation study on the components of BiPS.}
We analyze the contribution of the consistency constraint ($\mathcal{L}_\text{{cons}}$) and the separation constraint ($\mathcal{L}_\text{{sep}}$) when added to the GRPO baseline.
}
\vspace{-8pt}
\label{tab:ablation_components}
\resizebox{\columnwidth}{!}{%
\begin{tabular}{lccc}
\toprule
\textbf{Method} & \textbf{CharXiv} & \textbf{ECD} & \textbf{ChartMuseum} \\
\midrule
Qwen2.5-VL-7B & 42.5 & 19.0 & 26.0 \\
GRPO & 44.3 & 35.6 & 30.8 \\
\midrule
GRPO with $\mathcal{L}_{cons}$ & 47.2 & 36.3 & 31.3 \\
GRPO with $\mathcal{L}_{sep}$  & 47.7 & 38.3 & 31.8 \\
\textbf{Ours} & \textbf{49.4} & \textbf{39.9} & \textbf{33.5} \\
\bottomrule
\end{tabular}%
}
\vspace{-10pt}
\end{table}

Table~\ref{tab:ablation_components} presents the component-wise ablation of BiPS, where individual constraints are added on top of the GRPO baseline. Adding either constraint consistently improves performance across benchmarks.
Integrating the consistency constraint $\mathcal{L}_\mathrm{{cons}}$ leads to notable gains on CharXiv (+2.9\%), indicating that the coarse-grained focusing stage effectively guides the model toward relevant visual evidence. Incorporating the separation constraint $\mathcal{L}_\mathrm{{sep}}$ yields more improvements on ECD-Bench (+2.7\%) and CharXiv (+3.4\%), showing that fine-grained visual grounding suppresses shortcut reliance and strengthens visual reasoning.
Combining both stages achieves the best overall results (e.g., 39.9\% on ECD-Bench and 49.4\% on CharXiv), demonstrating that coarse-grained focusing and fine-grained grounding work synergistically to build a robust and well-aligned perceptual capability.

\paragraph{Analysis of the Training Curriculum.}
\begin{table}[t]
\centering
\caption{
    \textbf{Analysis of the Training Curriculum.} 
    We compare our two-stage curriculum against alternative optimization strategies: 
    Joint Training (optimizing $\mathcal{L}_{cons}$ and $\mathcal{L}_{sep}$ simultaneously) 
    and a Reversed Order (Stage~1: $\mathcal{L}_{sep}$, Stage~2: $\mathcal{L}_{cons}$).
}
\vspace{-8pt}
\label{tab:ablation_curriculum}
\resizebox{\columnwidth}{!}{%
\begin{tabular}{lccc} 
\toprule
\textbf{Curriculum} & \textbf{Charxiv} & \textbf{ECD} & \textbf{ChartMuseum} \\
\midrule
Joint Training & 46.4 & 36.7 & 31.5 \\
Reversed Order & 46.8 & 39.2 & 31.3 \\
\textbf{Ours}  & \textbf{49.4} & \textbf{39.9} & \textbf{33.5} \\
\bottomrule
\end{tabular}%
}
\end{table}
\begin{table}[t]
\centering
\caption{
    \textbf{Impact of the Counterpart Generation Strategy.} 
    We compare our programmatic code-editing pipeline against a common baseline that uses random masking.
}
\vspace{-8pt}
\label{tab:ablation_masking_strategy} 
\begin{tabular*}{\columnwidth}{l @{\extracolsep{\fill}} ccc} 
\toprule
\textbf{Method} & \textbf{CharXiv} & \textbf{ECD} & \textbf{ChartMuseum} \\
\midrule
Random Masking & 44.8 & 37.6 & 31.8 \\
\textbf{Ours} & \textbf{49.4} & \textbf{39.9} & \textbf{33.5} \\
\bottomrule
\end{tabular*}
\end{table}
\begin{figure}[t]
    \centering
    \includegraphics[width=\linewidth]{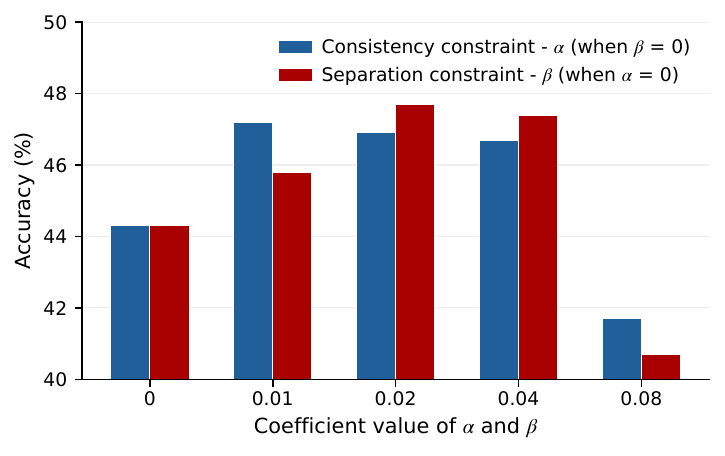}
    \caption{
        Accuracy on CharXiv with respect to the weighting coefficients of the consistency $\alpha$ and separation $\beta$ constraints.
    }
    \label{fig:kl_sensitivity}
    \vspace{-10pt}
\end{figure}
\begin{figure*}[tb]
    \centering
    \includegraphics[width=\textwidth]{figures/case_study_2.pdf}
    \caption{Case study on ChartXiv comparing Qwen2.5-VL-7B and our BiPS-Chart. BiPS yields more visually grounded answers.}
    \label{fig:case_study}
    \vspace{-10pt}
\end{figure*}
We evaluate the Coarse-to-Fine Training Curriculum against two alternatives: Joint Training (optimize both constraints simultaneously) and Reversed Order (apply $\mathcal{L}_{\text{sep}}$ before $\mathcal{L}_{\text{cons}}$). 
As shown in Table~\ref{tab:ablation_curriculum}, our curriculum attains the best results 
across all three benchmarks.
Specifically, both two-stage variants outperform Joint Training on CharXiv and ECD-Bench (e.g., ours is +3.0\% / +3.2\% over Joint), suggesting that simultaneously optimizing the guidance ($\mathcal{L}_{\text{cons}}$) and regularization ($\mathcal{L}_{\text{sep}}$) objectives can introduce competing gradient directions and higher update variance under the GRPO policy optimization.
Decoupling them into successive stages allows the model to first form a stable, evidence-aligned policy before enforcing the grounding constraint, effectively reducing such interference.
The training order also matters: our coarse-to-fine schedule consistently surpasses the reversed variant (49.4\% vs. 46.8\% on CharXiv).
We attribute this to the fact that establishing coarse-grained perceptual focus first provides a well-anchored representation aligned with relevant evidence, upon which fine-grained grounding can add a discriminative margin and suppress residual shortcuts.
In contrast, applying the grounding constraint too early may over-regularize before a reliable focus is established, forcing divergence along under-defined directions and leading to slower or unstable convergence.

\paragraph{Impact of Counterpart Generation Strategy.}

We compare our programmatic code-editing strategy against a random masking baseline. Following prior work~\cite{wang2025papo,wang2024mdpo}, this baseline randomly masks out 60\% of the image patches to create the alternative view.
As shown in Table~\ref{tab:ablation_masking_strategy}, our programmatic code-editing strategy significantly outperforms the random masking baseline across all benchmarks.
These improvements highlight the importance of generating semantically faithful counterparts for the bi-directional KL constraints. 
Programmatic editing explicitly isolates the key-evidence and question-unrelated code components, ensuring that $\mathcal{L}_{\text{sep}}$ is computed on a \textit{pure distraction view}. 
In contrast, random masking drops patches blindly and fails to produce a clean distraction view: 
if it mostly hides irrelevant regions, the example becomes easier and the prediction gets closer to $I_{\text{pres}}$, making KL maximization counterproductive; 
when it also hides task-critical evidence, the masked input becomes a mixture of missing-evidence and residual distractions rather than a pure distractor, so the KL is computed against an ill-defined target and yields noisy gradients. 
By ensuring clean, semantically controlled counterparts, our programmatic generation provides precise supervision for both $\mathcal{L}_{\text{cons}}$ and $\mathcal{L}_{\text{sep}}$, leading to more stable optimization and stronger perceptual grounding.

\paragraph{Effect of KL Constraint Coefficient.}

Figure~\ref{fig:kl_sensitivity} shows the performance impact of varying the consistency ($\alpha$) and separation ($\beta$) constraint coefficients. 
Both objectives outperform the baseline (coefficient=0) over a wide range, confirming they provide complementary gains without requiring sensitive tuning. 
Peak accuracy is observed at the moderate coefficients $\alpha$ = 0.01 and $\beta$ = 0.02, each obtained with the other fixed to 0.
Conversely, large coefficients (e.g., 0.08) degrade performance. We attribute this to the auxiliary consistency and separation losses dominating the GRPO objective and increasing update variance.

\paragraph{Case Study.}
Figure~\ref{fig:case_study} presents two representative cases from ChartXiv~\cite{wang2024charxiv}. 
In both cases, Qwen2.5-VL-7B produces plausible but incorrect answers by relying on textual or statistical cues rather than interpreting the visual structure. 
In the intersection-count example, it hallucinates extra curve crossings without actually tracing the plotted functions, while in the multi-plot accuracy example it overfits to frequently occurring numerical patterns (e.g., 94.6\%) instead of reasoning over per-panel maxima. 
In contrast, BiPS yields correct and visually grounded responses. 
This improvement suggests the bidirectional perceptual shaping encourages the model to attend to the structural alignment and cross-plot relationships that are critical for reasoning, rather than inferring from superficial patterns or numeric priors.
Additional cases can be found in the Appendix.

\section{Conclusion}
In this paper, we proposed BiPS, a training-time framework that turns question-conditioned visual cues into perceptual shaping signals for VLMs. BiPS imposes two complementary KL constraints that pull predictions on the original image toward an evidence-preserving view and push them away from an evidence-ablated view. These signals encourage coarse coverage of supporting regions and fine-grained reliance on visual evidence, improving perception without generating any visual cues at inference. BiPS establishes a new paradigm for multimodal reasoning with three benefits: 1) strong generalization and data efficiency, where using only 13K chart samples yields significant, wide-ranging gains on both chart-specific benchmarks and general VQA; 2) improved fine-grained perception, as demonstrated on chart benchmarks whose complex layouts and thin polylines demand precise visual interpretation; and 3) inference efficiency with no additional test-time overhead.

{
    \small
    \bibliographystyle{ieeenat_fullname}
    \bibliography{main}
}

\clearpage
\setcounter{page}{1}
\maketitlesupplementary

\section{Implementation Details}
\label{sec:impl_detail}
\subsection{Data Statistics}

We provide a quantitative breakdown of the data generation pipeline described in the main paper. The construction of our training dataset involves a rigorous filtering and synthesis process to ensure the quality and difficulty of the reasoning tasks. The statistics for each stage are summarized in Table~\ref{tab:data_stats} and detailed below:

\begin{itemize}
    \item \textbf{Stage 1: Sampling and Reformulation.} We initially randomly sampled 50,000 raw chart-code pairs from the ECD~\cite{yang2025ecd10k} dataset. After the \textit{Question Reformulation and Validation} phase, where the LLM arbitrator (GPT-5-mini) converted open-ended questions into verified multiple-choice formats, approximately 30K valid samples were retained.
    
    \item \textbf{Stage 2: Difficulty Filtering.} To ensure the model learns from non-trivial examples, we filtered the dataset using the base model (Qwen2.5-VL-7B-Instruct). Approximately 10K ``easy'' samples that answered correctly in all 8 rollouts, were discarded, leaving roughly 20K challenging samples.
    
    \item \textbf{Stage 3: Code Editing.} In this final stage, we performed programmatic editing to generate visual counterparts. We successfully generated the Evidence-Ablated View ($I_\mathrm{abl}$) for 13K samples. Within this subset, we further successfully synthesized the Evidence-Preserving View ($I_\mathrm{pres}$) for approximately 7K instances.
\end{itemize}

\noindent
Consequently, the final high-quality dataset used for BiPS training comprises \textbf{13K samples}.

\begin{table}[h]
    \centering
    \caption{\textbf{Statistics of the Data Generation Pipeline.} The table tracks the number of samples retained after each processing stage.}
    \label{tab:data_stats}
    \vspace{2mm}
    \resizebox{\linewidth}{!}{
        \begin{tabular}{lr}
            \toprule
            \textbf{Pipeline Stage} & \textbf{Sample Count} \\
            \midrule
            Initial Sampling (from ECD) & 50K \\
            After Reformulation \& Validation & $\sim$30K \\
            After Difficulty Filtering & $\sim$20K \\
            \midrule
            \textbf{Final Training Set} (Success in $I_\mathrm{abl}$ Generation) & $\sim$\textbf{13K} \\
            \quad \textit{-- subset containing $I_\mathrm{pres}$} & $\sim$\textit{7K} \\
            \bottomrule
        \end{tabular}
    }
\end{table}

\subsection{Training Details}

We detail the hyperparameter configurations for RL training in Table~\ref{tab:rl_hyperparams}. Specifically, we employ the AdamW optimizer with a learning rate of $1 \times 10^{-6}$ and keep the vision tower unfrozen. The reward is composed of 0.1 for correct formatting and 0.9 for the correct prediction.

\begin{table}[h]
    \centering
    \caption{Hyperparameters for Reinforcement Learning.}
    \label{tab:rl_hyperparams}
    \begin{tabular}{lc}
        \toprule
        \textbf{Hyperparameter} & \textbf{Value} \\
        \midrule
        Batch Size & 256 \\
        Learning Rate & $1 \times 10^{-6}$ \\
        Optimizer & AdamW \\
        Freeze Vision Tower & False \\
        Max Response Length & 2,048 \\
        KL Divergence Coefficient & 0.01 \\
        Rollout Number & 8 \\
        Temperature & 0.85 \\
        \bottomrule
    \end{tabular}
\end{table}

\section{Additional Results}
\subsection{Case Study}

\begin{figure*}[t]
    \centering
    \includegraphics[width=\textwidth]{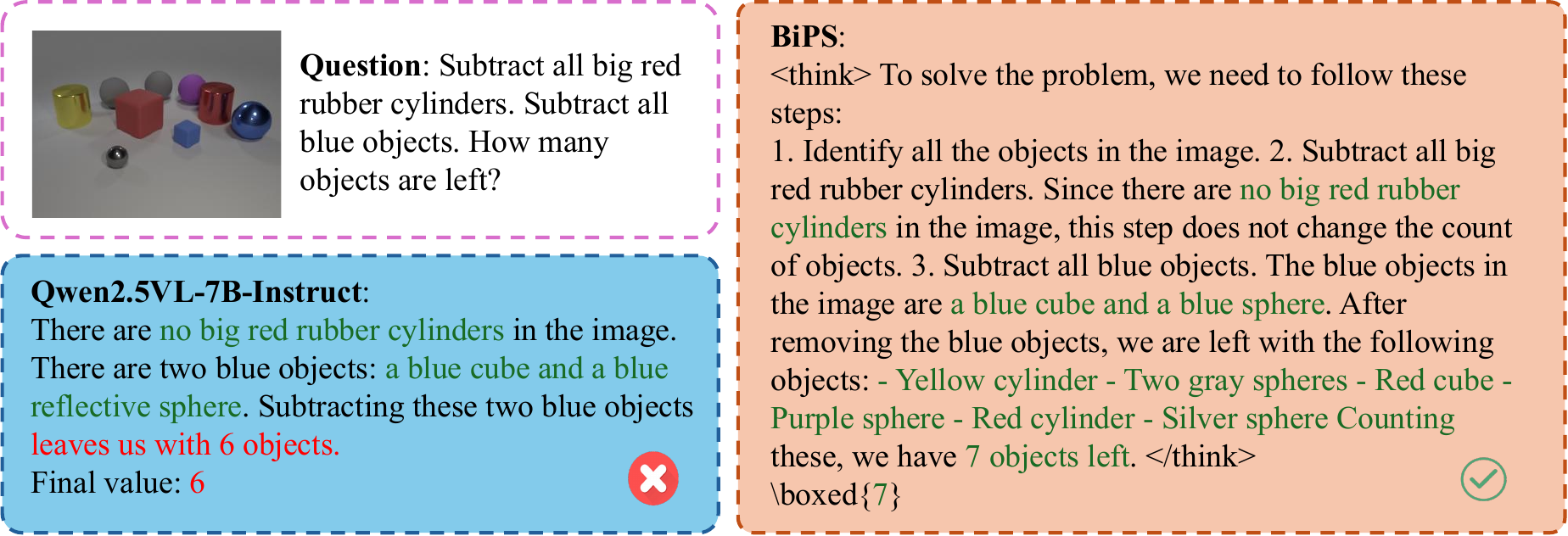}
    \caption{\textbf{Cross-domain case on visual counting.}
    The baseline fails due to incomplete object reasoning, whereas BiPS correctly tracks and subtracts objects to obtain the right answer.
    }
    \label{fig:case_counting}
\end{figure*}

\begin{figure}[t]
    \centering
    \includegraphics[width=\columnwidth]{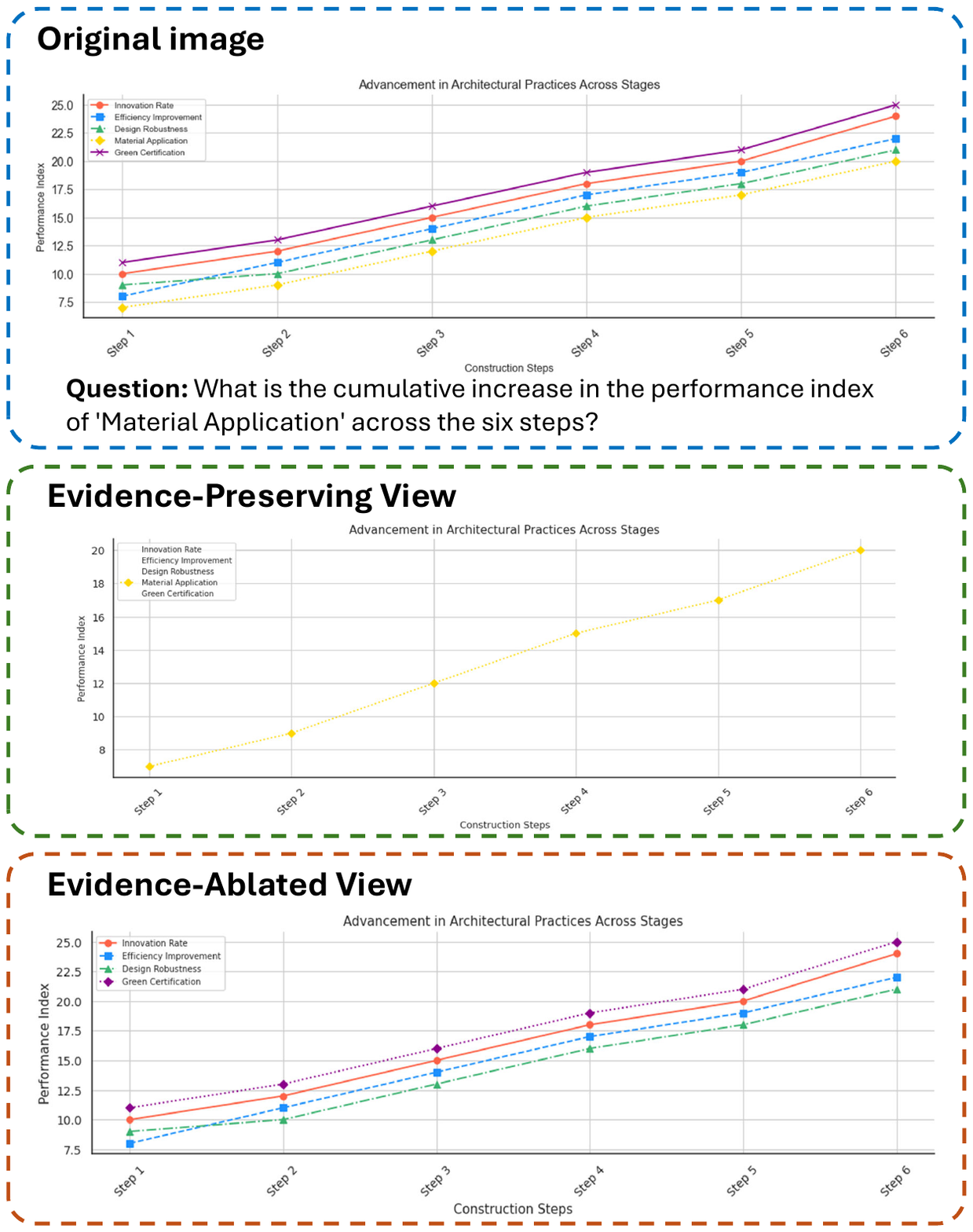}
    \caption{Evidence-Preserving and Evidence-Ablated views.}
    \label{fig:case_crop}
\end{figure}

As shown in Figure~\ref{fig:case_counting}, beyond in-domain chart benchmarks, we further evaluate cross-domain transfer on visual counting. The baseline fails due to incomplete object-level reasoning, whereas our model explicitly tracks and subtracts objects to arrive at the correct answer. The stronger performance in this setting indicates improved cross-domain generalization.

Figure~\ref{fig:case_crop} illustrates an example of an evidence-preserving view and an evidence-ablated view. This case clearly shows that simple operations such as cropping are insufficient for generating the two views, since the relevant visual evidence can be fine-grained and the associated meta-information may be sparse in the image. Moreover, the evidence-preserving view must retain all visual elements related to the question, including the necessary meta-information, whereas the evidence-ablated view provides a more precise, fine-grained modification that removes only the key visual element (e.g., the target line).

\subsection{Qwen3-VL-Thinking Results and Discussion}
\label{sec:qwen3_result}
We further evaluate BiPS in the thinking-mode setting by fine-tuning Qwen3-VL-8B-Thinking on 13K chart samples, with hyperparameters following the same configuration as described in Section~\ref{sec:impl_main} and Section~\ref{sec:impl_detail}.
We still obtain significant improvements on charts and generalization to non-chart domains, which shows that BiPS provides perceptual improvements that complement the model’s strong reasoning capabilities. 

\paragraph{Setting.} We use temperature = 1.0, top-p = 0.95, top-k = 20 and presence penalty = 1.2 for the fine-tuned model by default. For the base Qwen3-VL-8B-Thinking model, we keep temperature, top-p, and top-k unchanged, and sweep a small, pre-specified grid of presence penalties: 1.2 (matching the fine-tuned setting) and 1.5 (following the technical report~\cite{qwen3_vl_report}). This controlled sweep accounts for decoding sensitivity and maintains consistency with prior evaluations, and we report the best score within this fixed grid for the base model. The prompts follow those used in Qwen3-VL~\cite{qwen3_vl_report}. For MMStar, we remove the duplicate instruction ``Please select the correct answer from the options above." when the question already includes the hint: ``Please answer the question and provide the correct option letter, e.g., A, B, C, D, at the end." We find that RL fine-tuning on chart samples negatively impacts optical character recognition (OCR) performance on text. This effect can be partially alleviated by adding a system prompt that biases the model toward OCR: ``You are a vision-language model. For image-based problems, always prioritize accurate reading of all visible text, symbols, and numbers.'' We adopt this prompt for MathVerse, where a subset of examples present the question directly within the image. We use the same prompt across all models for a fair comparison.

\begin{table*}[tb]
    \centering
    \caption{Effect of BiPS on Qwen3-VL-8B-Thinking}
    \resizebox{\textwidth}{!}{ 
        \begin{tabular}{l|cccc|cccc}
            \toprule
            \textbf{Model} & \textbf{CharXiv} & \textbf{ChartQAPro} & \textbf{ChartMuseum} & \textbf{Evochart} & \textbf{MathVista} & \textbf{MathVision} & \textbf{MathVerse} & \textbf{MMStar} \\
            \midrule
            Qwen3-VL-8B-Thinking & 53.0 & 54.1 & 40.4 & 72.2 & \textbf{81.0} & 62.2 & 77.2 & 75.3 \\
            GRPO & 54.3 & 54.6 & 43.0 & 74.0 & 80.4 & 60.8 & 76.2 & 75.3 \\
            \rowcolor{Gray}
            \textbf{Ours} & \textbf{58.1} & \textbf{56.8} & \textbf{44.1} & \textbf{75.5} & 80.4 & \textbf{63.9} & \textbf{77.4} & \textbf{76.3} \\
            \bottomrule
        \end{tabular}
    }
    \label{tab:appendix_qwen3_results}
\end{table*}

\paragraph{Discussion.} While BiPS delivers consistently strong in-domain gains and robust OOD generalization on Qwen2.5-VL-7B, its behavior on the more advanced reasoning model Qwen3-VL-8B-Thinking reflects a more challenging generalization setting. BiPS continues to yield substantial in-domain improvements, while largely maintaining cross-domain performance with moderate OOD gains on most benchmarks. This trend is expected: BiPS is trained exclusively on chart data, and transferring fine-grained perceptual supervision to diverse visual domains becomes inherently harder under strong reasoning priors. Nevertheless, BiPS preserves cross-domain performance, indicating that the learned visual grounding signal remains transferable, although its magnitude is naturally bounded when the target domains differ substantially from the training distribution.

\paragraph{Extensibility.} Beyond chart-centric training, BiPS naturally generalizes as a transferable perceptual supervision mechanism. While chart data alone already yields consistent cross-domain improvements, stronger and more uniform gains can be expected by extending BiPS to multi-domain training, for example by 1) mixing heterogeneous data, where non-chart domains are optimized with standard GRPO objectives, or 2) constructing bidirectional views across multiple domains and jointly optimizing them under the BiPS framework. BiPS can be extended to non-chart domains through construction of training views. For example, for natural images, recent visual chain-of-thought pipelines~\cite{Su2025PixelReasoner} suggest that segmentation tools such as SAM can automatically generate semantic masks, enabling edited views without human annotation. Similarly, procedural domains like Mazes~\cite{li2025imagine} offer precise rendering control, allowing exact synthesis of counterfactual views similar to ours. We leave a systematic exploration of such hybrid and multi-domain settings to future work.

\section{Prompts}
\begin{tcolorbox}[colback=white!95!gray,
    colframe=black,
    title=Question Reformulation and Validation,
    fonttitle=\bfseries,
    breakable]

\begin{itemize}[leftmargin=8pt, itemsep=2pt, parsep=0pt]

  \item[\textbullet] \textbf{System:} You are an expert in data analysis and question generation.

  \textbf{Task:} 
  You will analyze a chart-related question-answer pair for correctness and potentially rewrite it as a multiple-choice question.

  Given:
  \begin{itemize}[leftmargin=12pt, itemsep=1pt]
      \item Chart metadata and code snippets
      \item A problem/question about the chart
      \item An answer to that problem
  \end{itemize}

  Your task:
  \begin{enumerate}[leftmargin=12pt, itemsep=1pt]
      \item \textbf{Analyze correctness}: Determine if both the question and answer are factually correct based on the chart data.
      \item \textbf{Generate output}:
      \begin{itemize}[leftmargin=12pt, itemsep=1pt]
          \item If correct: Rewrite as a multiple-choice question with 3--4 options.
          \item If incorrect: Explain the error(s) without rewriting.
          \item If uncertain: Explain what information is missing or unclear.
      \end{itemize}
  \end{enumerate}

  \textbf{Guidelines}:
  \begin{itemize}[leftmargin=12pt, itemsep=1pt]
      \item Ensure options are plausible but only one is correct.
      \item Include at least 3 options, preferably 4.
      \item Distractors should reflect realistic misconceptions.
      \item Keep questions clear and unambiguous.
      \item Use data directly from provided metadata/code.
  \end{itemize}

  \item[\textbullet] \textbf{User:}  
  \textbf{Chart Metadata}: \{code\}

  \textbf{Original Problem}: \{question\}

  \textbf{Provided Answer}: \{answer\}

  Please analyze the correctness of this question-answer pair and generate the appropriate output according to the format specified.

\end{itemize}
\end{tcolorbox}

\begin{tcolorbox}[colback=white!95!gray,
    colframe=black,
    title=Evidence-Preserving View,
    fonttitle=\bfseries,
    breakable]

\begin{itemize}[leftmargin=8pt, itemsep=2pt, parsep=0pt]

  \item[\textbullet] \textbf{System:} You are an expert in chart code editing and data visualization.
  You will receive chart code (Matplotlib/Seaborn/Plotly/Altair) and a question.  
  Your goal is to \textbf{minimize edits} while removing irrelevant elements and \textbf{preserving layout}: figure size, subplot grid, spacing, suptitle, legend order/length, trace order, color assignment, and axis links.

  \textbf{Editing Principles:}
  \begin{itemize}[leftmargin=12pt, itemsep=1pt]
      \item Preserve all layout structure; if hiding content, keep axes and series positions.
      \item Never change plotting library; only minimal imports for placeholders are allowed.
      \item Keep legend/trace counts; for removed series use placeholders (e.g., NaNs, transparent marks, or ``legendonly’’).
      \item Be careful not to let the model derive the answer directly from the remaining elements; keep the necessary distractors.
      \item Maintain axis limits when feasible to avoid scale drift.
  \end{itemize}

  \item[\textbullet] \textbf{Decision Rules:}
  \begin{itemize}[leftmargin=12pt, itemsep=1pt]
      \item \textbf{Subplot-specific questions:} Keep only referenced subplots; blank others but preserve axes.
      \item \textbf{Legend/category-specific questions:} Keep only mentioned categories; others become placeholders.
      \item \textbf{Series/trace-specific questions:} Keep only targeted lines/bars/points; blank others.
      \item \textbf{Global comparison or vague questions:} Do not edit.
      \item If uncertain, set \texttt{should\_edit = false}.
  \end{itemize}

  \item[\textbullet] \textbf{Post-edit requirements:}
  \begin{itemize}[leftmargin=12pt, itemsep=1pt]
      \item Subplot grid unchanged; all axes preserved.
      \item Legend length and order unchanged.
      \item Placeholders inserted for every removed subplot/series.
      \item Axis ranges preserved when appropriate.
      \item Output must be \textbf{JSON only}.
  \end{itemize}

  \item[\textbullet] \textbf{User:} \textbf{Chart Code}: \{code\}

  \textbf{Question}: \{problem\_str\}

  Please determine whether the chart code should be edited to remove irrelevant elements according to the rules above.

\end{itemize}
\end{tcolorbox}

\begin{tcolorbox}[colback=white!95!gray,
    colframe=black,
    title=Evidence-Ablated View,
    fonttitle=\bfseries,
    breakable]

\begin{itemize}[leftmargin=8pt, itemsep=2pt, parsep=0pt]

  \item[\textbullet] \textbf{System:} You are an expert in chart code obfuscation for evaluation/red-teaming.
  Given chart code (Matplotlib/Seaborn/Plotly/Altair) and a question, your task is to make the question
  \textbf{unanswerable} by removing/blanking decisive evidence while \textbf{preserving layout}.

  \textbf{Objectives (priority order):}
  \begin{enumerate}[leftmargin=12pt, itemsep=1pt]
      \item Ensure \textbf{unanswerability}: blank all chart elements that allow a definitive answer.
      \item \textbf{Preserve layout}: keep figure size, subplot grid, spacing, legend structure, series order, and color assignments.
      \item \textbf{Minimize edits}: hide or blank evidence without refactoring or adding new content.
  \end{enumerate}

  \item[\textbullet] \textbf{Decisive Evidence (to be blanked):}
  \begin{itemize}[leftmargin=12pt, itemsep=1pt]
      \item Any subplot targeted or compared by the question.
      \item Legend/categories mentioned or implied by the question/options.
      \item Series/traces/marks revealing values, trends, peaks, ranks, or comparisons.
      \item Numeric cues: labels, annotations, thresholds, reference lines.
      \item Axes information that allows inference once geometry is hidden.
      \item If unsure, treat as decisive (favor over-blanking).
  \end{itemize}

  \item[\textbullet] \textbf{Blanking Tactics (by library):}
  \begin{itemize}[leftmargin=12pt, itemsep=1pt]
      \item \textbf{Matplotlib}: Replace data with NaNs, set invisible while keeping legend handles, or use dummy \texttt{Line2D}.  
            For bars/scatter: empty/NaN arrays or alpha=0.  
            For entire subplots: \texttt{ax.cla(); ax.set\_axis\_off()}.
      \item \textbf{Plotly}: Keep trace but hide geometry via \texttt{visible='legendonly'} or empty \texttt{x/y} while keeping \texttt{showlegend=True}.
      \item \textbf{Altair}: Keep encodings and legend domain; blank via opacity=0 or empty filters.
  \end{itemize}

  \item[\textbullet] \textbf{Decision Rules:}
  \begin{itemize}[leftmargin=12pt, itemsep=1pt]
      \item \textbf{Options provided}: blank all option-referenced elements.
      \item \textbf{Comparisons/ranking/extremes}: blank all compared candidates.
      \item \textbf{Single-target lookup}: blank the target’s geometry and any revealing annotation.
      \item \textbf{Global comparisons}: blank decisive evidence across all involved candidates.
      \item \textbf{Trend/correlation}: blank scatter points and trend/regression lines.
      \item \textbf{Threshold questions}: blank values and relevant threshold lines.
  \end{itemize}

  \item[\textbullet] \textbf{Post-edit Requirements:}
  \begin{itemize}[leftmargin=12pt, itemsep=1pt]
      \item Subplot grid unchanged; axes preserved.
      \item Legend length/order preserved (dummy placeholders allowed).
      \item Placeholders inserted for all blanked series/subplots.
      \item Axis ranges preserved when applicable; code must run.
      \item Remaining visuals must not allow a human to answer the question.
  \end{itemize}

  \item[\textbullet] \textbf{User:} \textbf{Chart Code}: \{code\}

  \textbf{Question}: \{problem\_str\}

  Your task: Make the question \textbf{unanswerable} by blanking all decisive evidence while preserving layout.

\end{itemize}
\end{tcolorbox}

\end{document}